\documentclass[sigconf]{acmart}
\usepackage{multirow}
\usepackage{booktabs}       
\usepackage{pifont}
\usepackage{graphicx}       
\usepackage{wrapfig}
\usepackage{dsfont}
\usepackage{subcaption}
\usepackage{algorithm}
\usepackage{algorithmic}
\usepackage{natbib}
\usepackage{url}  
\usepackage{hyperref}

\definecolor{hlred}{HTML}{FFDDDD}
\definecolor{hlorange}{HTML}{FFEECC}

\newcommand{\Max}[1]{\cellcolor{hlred}#1}
\newcommand{\Second}[1]{\cellcolor{hlorange}#1}

\usepackage[textsize=tiny]{todonotes}

\usepackage{amsfonts}
\usepackage{xcolor}
\usepackage[table]{xcolor}
\usepackage{dsfont}
\hypersetup{
    colorlinks=true,
    linkcolor=blue,
    citecolor=blue,
    urlcolor=blue,
    breaklinks=true
}

\AtBeginDocument{%
  }

\usepackage[most]{tcolorbox}
\usepackage{xcolor}

\newtcolorbox{discoverybox}{
  enhanced,
  colback=black!3,
  colframe=black!35,
  boxrule=0.4pt,
  arc=1mm,
  left=1.2mm,
  right=1.2mm,
  top=0.8mm,
  bottom=0.8mm,
  before skip=4pt,
  after skip=4pt
}

\newtcolorbox{corollarybox}{
  enhanced,
  colback=green!4,
  colframe=green!35!black,
  boxrule=0.45pt,
  arc=1mm,
  left=1.2mm,
  right=1.2mm,
  top=0.8mm,
  bottom=0.8mm,
  before skip=4pt,
  after skip=4pt
}

\copyrightyear{2026}
\acmYear{2026}
\setcopyright{cc}
\setcctype{by-nc-nd}
\acmConference[MM '26]{Proceedings of the 34th ACM International Conference on Multimedia}{November 10--14, 2026}{Rio de Janeiro, Brazil}
\acmBooktitle{Proceedings of the 34th ACM International Conference on Multimedia (MM '26), November 10--14, 2026, Rio de Janeiro, Brazil}
\acmDOI{10.1145/3767308.3835698}
\acmISBN{979-8-4007-2213-4/2026/11}

\begin{document}

\title{SplitGaussian: Reconstructing Dynamic Scenes via Visual Geometry Decomposition}

%
\author{Lechao Cheng}
\email{chenglc@hfut.edu.cn}
\affiliation{%
  \institution{Hefei University of Technology}
  \city{Hefei}
  \state{Anhui}
  \country{China}
}
\author{Jiahui Li}
\email{22321277@zju.edu.cn}
\affiliation{%
  \institution{Zhejiang University}
  \city{Hangzhou}
  \state{Zhejiang}
  \country{China}
}

\author{Jingxuan He}
\email{jihe0215@uni.sydney.edu.au}
\affiliation{%
  \institution{University of Sydney}
  \city{Sydney}
  \country{Australia}
}

\author{Shengeng Tang}
\email{tangsg@hfut.edu.cn}
\affiliation{%
  \institution{Hefei University of Technology}
  \city{Hefei}
  \state{Anhui}
  \country{China}
}

\author{Gang Huang}
\email{huanggang@zju.edu.cn}
\affiliation{%
  \institution{Zhejiang University}
  \city{Hangzhou}
  \state{Zhejiang}
  \country{China}
}

\author{Tianrui Hui}
\email{huitianrui@gmail.com}
\affiliation{%
  \institution{Hefei University of Technology}
  \city{Hefei}
  \state{Anhui}
  \country{China}
}
\author{Yaxiong Wang}
\email{wangyx@hfut.edu.cn}
\affiliation{%
  \institution{Hefei University of Technology}
  \city{Hefei}
  \state{Anhui}
  \country{China}
}

\author{Zhun Zhong}
\authornotemark[1]
\email{zhunzhong007@gmail.com}
\affiliation{%
  \institution{Hefei University of Technology}
  \city{Hefei}
  \state{Anhui}
  \country{China}
}








\renewcommand{\shortauthors}{Lechao Cheng et al.}

\begin{abstract}
  Reconstructing dynamic 3D scenes from monocular video remains fundamentally challenging due to the need to jointly infer motion, structure, and appearance from limited observations. Existing dynamic scene reconstruction methods based on Gaussian Splatting often entangle static and dynamic elements in a shared representation, leading to motion leakage, geometric distortions, and temporal flickering. We identify that the root cause lies in the coupled modeling of geometry and appearance across time, which hampers both stability and interpretability. To address this, we propose \textbf{SplitGaussian}, a novel framework that explicitly decomposes scene representations into static and dynamic components. By decoupling motion modeling from background geometry and allowing only the dynamic branch to deform over time, our method prevents motion artifacts in static regions while supporting view- and time-dependent appearance refinement. This disentangled design not only enhances temporal consistency and reconstruction fidelity but also accelerates convergence. Extensive experiments demonstrate that our approach outperforms prior state-of-the-art methods in rendering quality, geometric stability, and motion separation.
\end{abstract}

\ccsdesc[500]{Computing methodologies~Reconstruction}



\keywords{Dynamic Scene Reconstruction, 3DGS}


\maketitle

\begin{figure}[!t]
    \centering
    \includegraphics[width=\linewidth]{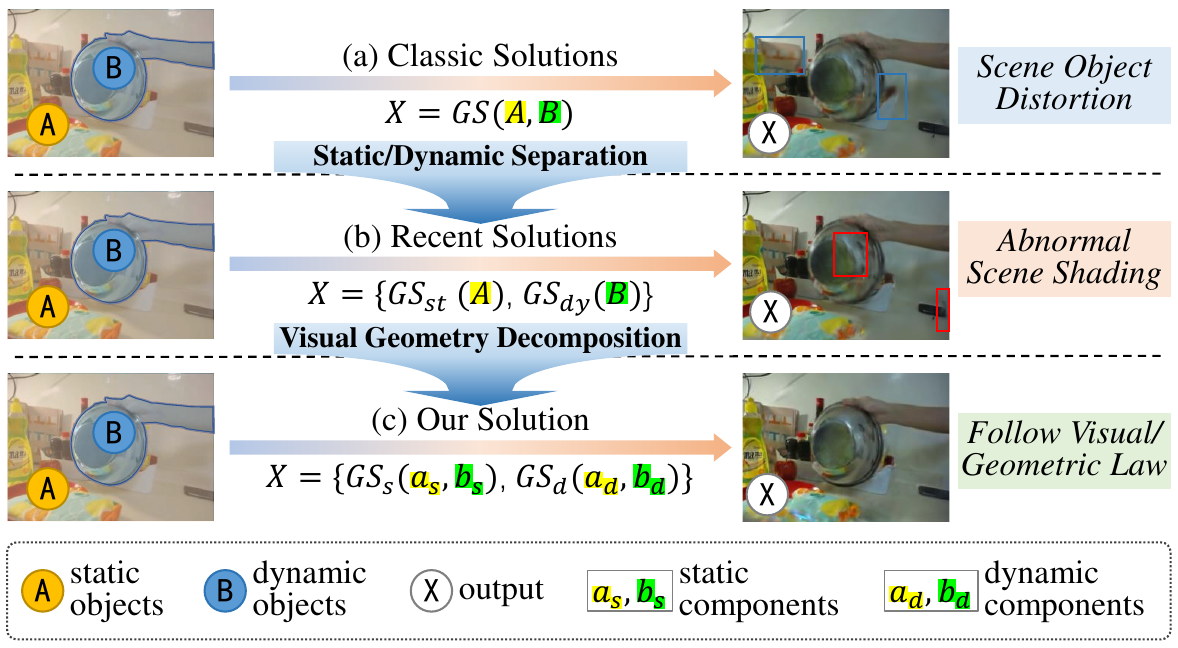}
    \vspace{-7mm}
\caption{\small \textbf{Motivation.} 
    (a) Classical monocular dynamic reconstruction entangles motion and background, leading to geometric distortions and shading artifacts. 
    (b) Recent dynamic 3DGS methods still rely on deformation fields, causing motion leakage and temporal flicker. 
    (c) Our Visual Geometry Decomposition (VGD), which splits Gaussian primitives into static and dynamic attributions, yields temporally consistent and geometrically stable monocular reconstructions. }
    \label{fig:teaser}
    \vspace{-6mm}
\end{figure}

\section{Introduction}\label{sec:intro}
Reconstructing dynamic 3D scenes from monocular video remains a core challenge in computer vision, with far-reaching applications in virtual reality, free-viewpoint rendering, and autonomous perception~\cite{li2024gpnerf}.
These scenes present non-rigid motion, occlusions, and appearance variation, demanding joint inference of structure, motion, and camera pose from limited visual cues. Traditional multi-view stereo and depth sensors offer stronger constraints but restrict flexibility. Methods like NR‑NeRF~\cite{tretschk2021non} introduce a canonical volume plus deformation field to enable dynamic reconstruction from monocular video, but they require expensive per-scene ray-based optimization and converge slowly.

Compared to implicit volumetric fields, 3D Gaussian Splatting (3DGS)~\cite{kerbl3Dgaussians} offers explicit and compact representations, enabling faster optimization and real-time rendering. Early attempts at dynamic reconstruction with 3DGS treat each frame independently by reconstructing a separate set of Gaussians per frame~\cite{dynamic3dgs}, which fails to establish temporal consistency and leads to redundant or unstable representations. To this end, existing cutting-edge dynamic extensions, such as Deformable 3DGS~\cite{yang2023deformable3dgs}, use a unified deformation network across static and dynamic regions, often disturbing static structures and introducing temporal artifacts. When a single deformation field is applied uniformly across both dynamic and static regions, it tends to propagate motion artifacts into static areas, leading to geometric distortions (e.g., \textit{rigid structures slightly shifted or warped}) and appearance inconsistencies (e.g., \textit{temporal texture flicker or color drift}). This issue also persists in prior works~\cite{Wu_2024_CVPR, bard_gs, modecgs, compactgs}, where static scene elements exhibit lingering artifacts despite motion modeling.

Let us revisit the Gaussian Splatting process, where each primitive jointly encodes \textbf{visual appearance} (e.g., color defined by spherical harmonic coefficients, opacity) and \textbf{geometry} (e.g., position, rotation, and scaling). We argue that such joint modeling underlies many common artifacts in dynamic reconstruction, such as motion leakage into static regions and temporal inconsistencies. Existing methods like DeGauss~\cite{DeGauss} (depth-aware compositing), DynaSplat~\cite{dynasplat} (optical flow guidance), and GauFre~\cite{gaufre} (multi-stream processing) attempt to address dynamics, but all suffer from effectively disentangling dynamic and static regions from the underlying meta-representations. 

Specifically, real-world scenes are commonly rendered with global illumination: {\textit{{\textbf{Note (I)}} the motion of dynamic objects changes indirect lighting, cast shadows, and inter-reflections on nearby static surfaces, causing even strictly static geometry to exhibit time-varying radiation.}} Methods that enforce fixed appearance for static gaussian primitives cannot capture such effect, producing flicker or inconsistent shading. We therefore introduce a residual appearance model for the static branch, allowing appearance to vary over time while geometry remains fixed. 
By contrast, dynamic Gaussians require no explicit time-varying appearance; \textit{{\textbf{Note (II)}} their evolving geometry suffices to explain motion-induced illumination.
Adding dynamic appearance residuals creates ambiguity between deformation and appearance, destabilizes motion learning, and often re-introduces flicker.} Hence, explicit appearance modeling is applied only to static Gaussians, whereas dynamics rely on deformation-driven updates to explain appearance changes.

Motivated by this, we propose to decompose the visual geometry within the gaussian representation to explicitly model dynamic and static regions. Specifically, the static part maintains fixed geometry but allows appearance to vary over time, while the dynamic part models time-varying geometry and appearance through a deformation network conditioned on shared spatiotemporal encodings. This design effectively decouples motion modeling from background representation, yielding more robust and interpretable reconstructions as illustrated in Figure~\ref{fig:teaser}. We further improve reconstruction by applying visibility-driven pruning to remove low-contribution static Gaussians and introducing a depth-aware pretraining phase for better geometric initialization and depth consistency.
We summarize the key contributions as follows:

\begin{itemize}
    \item We introduce an explicit decomposition of Gaussian primitives into static and dynamic components, enabling disentangled modeling of geometry and appearance to improve reconstruction stability.
    \item We orchestrate a unified framework with shared spatiotemporal encoding, dedicated deformation network, and visibility-driven pruning for efficient and coherent dynamic scene reconstruction from monocular video.
    \item Extensive experiments demonstrate that our method achieves superior performance in reducing geometric distortions and appearance flickering, outperforming existing state-of-the-art baselines.
\end{itemize}

\section{Related Works}

\subsection{Dynamic Scene Reconstruction}
Dynamic scene reconstruction seeks to recover geometry and appearance under challenging conditions such as occlusions and illumination changes. Traditional methods, including multi-view stereo~\cite{combe} and scene flow~\cite{vogel2013piecewise}, require dense depth and struggle with significant deformations or fast motion. Learning-based approaches~\cite{ma2019deep} jointly predict geometry and motion from monocular inputs but often lack temporal coherence. Recent neural implicit methods~\cite{li2024nerfsurvey}, such as NeRF~\cite{mildenhall2020nerf}, utilize continuous volumetric representations with deformation fields (D-NeRF~\cite{pumarola2020d}, NSFF~\cite{li2020neural}) or higher-dimensional embeddings (HyperNeRF~\cite{park2021hypernerf}) for dynamic reconstruction. Despite their high visual fidelity, these methods involve costly per-scene optimization and slow inference, limiting real-time application. In contrast, 3DGS~\cite{kerbl3Dgaussians} employs rasterization-based anisotropic Gaussians, enabling efficient optimization and real-time rendering. Recent dynamic extensions~\cite{yang2023deformable3dgs, Wu_2024_CVPR, bard_gs, modecgs, dynasplat, lian2026dynamic3dgs} incorporate time-varying transformations into Gaussian to improve spatiotemporal consistency.

\subsection{Decomposition of Dynamic and Static Region}
Decomposing scenes into static and dynamic components simplifies motion modeling and enhances reconstruction quality~\cite{zou2025vdnerf}. Early methods such as DeGauss~\cite{DeGauss} and DynaSplat~\cite{dynasplat} employ external motion cues or optical flow-based masks, often using separate branches or losses. GauFre~\cite{gaufre} further introduces dual-branch architectures with occlusion reasoning to improve temporal coherence. CoGS~\cite{cogs} uses compositional modeling with learned masks for flexible blending, but at the cost of increased complexity. BARD-GS~\cite{bard_gs} introduces deformation modeling into static regions, addressing motion blur but risking static geometry distortion. Existing approaches like RoDyGS~\cite{rodygs} performs static–dynamic separation \emph{at the object level} through motion-basis learning and trajectory regularization, primarily aiming to stabilize long-range non-rigid motion. However, such a coarse separation strategy overlooks the dynamic interactions of radiance within a scene. In realistic scenarios~\cite{li2024loopgaussian}, the motion of dynamic objects often alters the \emph{local radiance distribution} of surrounding static backgrounds or nearby objects, whereas the reverse influence—from static regions to moving objects—is usually negligible. Consequently, an object-level “moving vs. static” decision fails to account for local illumination variations and radiance coupling, often leading to inconsistent appearance and lighting drift during reconstruction. We instead disentangle \emph{at the attribute level}: for \emph{static} regions, keep geometry fixed and strongly regularized while allowing color coefficients and opacity~$\alpha$ to adapt to \emph{local} radiance changes; for \emph{dynamic} regions, assume \emph{global} radiance consistency and focus capacity on modeling spatial motion states. By leveraging unified spatiotemporal encoding, visibility-driven pruning, and depth-aware pretraining, our representation-level design enables explicit disentanglement between geometry and appearance while preserving radiance consistency, without resorting to external segmentation or multi-encoder fusion. This effectively mitigates motion leakage and static artifacts, yielding sharper boundaries and more stable optimization dynamics.

\begin{figure*}
  \centering
  \includegraphics[width=\linewidth]{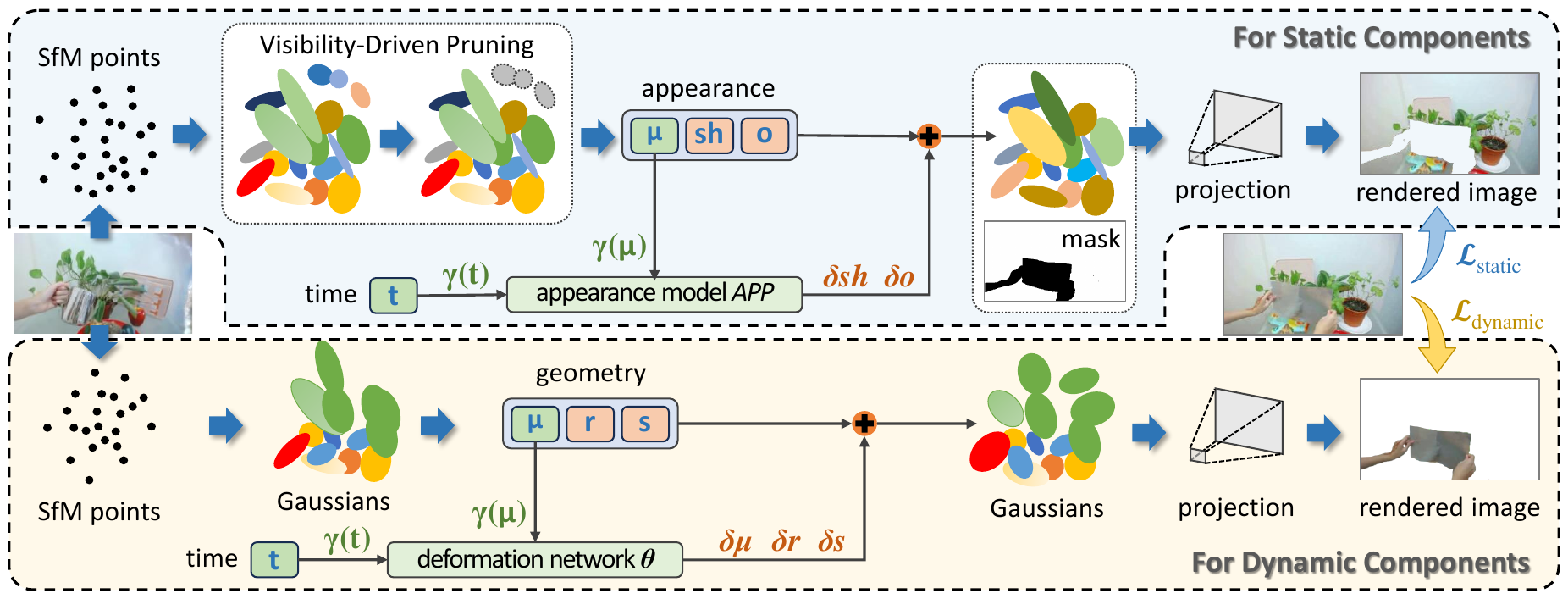}
  \vspace{-6mm}
  \caption{\small \textbf{Framework Overview}. We adopt a two-stage training pipeline: \textbf{Stage I} disentangles static and dynamic Gaussians via region-specific supervision and visibility-driven pruning to enhance geometric stability; \textbf{Stage II} jointly optimizes both components, where static appearance is modeled without deformation and dynamic motion is learned via a spatiotemporally-conditioned deformation network, enabling mutual refinement and improved reconstruction fidelity.}
  \label{fig:framework}
  \vspace{-3mm}
\end{figure*}

\section{Method}



We propose a dynamic scene reconstruction framework that explicitly decomposes geometry into static and dynamic components, which are modeled via a unified spatiotemporal encoding. Static Gaussians maintain fixed positions with time-varying appearance, while dynamic Gaussians undergo learned motion-based deformation. Reconstruction is supervised through region-specific losses guided by visibility masks, ensuring temporal consistency and disentanglement. A visibility-driven pruning strategy improves static reliability and efficiency, and depth-aware pretraining further refines geometry alignment. We will detail this later.
\subsection{Preliminary: 3D Gaussian Splatting}
3D Gaussian Splatting~\cite{kerbl3Dgaussians} represents a scene using a set of anisotropic Gaussians, each defined by a 3D center $\mu$, covariance matrix $\Sigma$, spherical harmonic (SH) color coefficients $C$, and opacity $\alpha$. The Gaussian density at a point $X$ is given by:
\begin{equation}
G(X) = \exp\left( -\frac{1}{2} X^\top \Sigma^{-1} X \right). \tag{1}
\end{equation}

For efficient optimization and interpretation, the covariance matrix is typically decomposed as:
\begin{equation}
\Sigma = R S S^\top R^\top, \tag{2}
\end{equation}
where $R$ is a rotation matrix and $S$ is a scaling matrix.

During rendering, the Gaussian is projected into screen space by applying the viewing transformation matrix $W$ and the Jacobian matrix $J$ of the affine approximation of the camera projection:
\begin{equation}
\Sigma' = J W \Sigma W^\top J^\top. \tag{3}
\end{equation}

The final pixel color is computed via alpha compositing in front-to-back order as:
\begin{equation}
C = \sum_{i=1}^{N} c_i \alpha_i \prod_{j=1}^{i-1} (1 - \alpha_j), \tag{4}
\end{equation}
where $c_i$ and $\alpha_i$ denote the color and opacity of the $i$-th Gaussian. This representation enables real-time rendering with high visual fidelity, but it inherently assumes static scene geometry. When extended to dynamic scenes, directly optimizing a shared set of Gaussians often introduces motion artifacts and temporal inconsistencies, due to the entanglement of geometry and appearance modeling. These challenges inspire our decomposition-based formulation, which explicitly separates static and dynamic components to enable more stable and physically interpretable reconstructions.

\subsection{Visual Geometry Decomposition}

A core challenge in dynamic scene reconstruction is simultaneously modeling time-varying geometry and appearance. To address this, we explicitly decompose the scene at time $t$ into two sets of Gaussian primitives:
\[
G(t): = \{G_{\mathrm{s}}(\mu_s, \Sigma_s, w_s(t))\} \cup \{G_{\mathrm{d}}(\mu_d(t), \Sigma_d(t), w_d)\}, \tag{5}
\]
where each Gaussian primitive $G(\mu, \Sigma, w)$ consists of:
\begin{itemize}
    \item \textbf{Appearance}: represented by attributes $w$, including spherical harmonic coefficients and opacity.
    \item \textbf{Geometry}: represented by center position $\mu$ and covariance matrix $\Sigma$.
\end{itemize}

In our formulation:
\begin{itemize}
    \item The static component $G_{\mathrm{s}}$ maintains fixed geometry ($\mu_s, \Sigma_s$) and only allows temporal variation in appearance $w_s(t)$.
    \item The dynamic component $G_{\mathrm{d}}$ models time-varying geometry 
      ($\mu_d(t), \Sigma_d(t)$) while keeping its appearance $w_d$ fixed.
\end{itemize}

This explicit separation of geometry and appearance preserves static background integrity and accurately isolates dynamic regions. 

\noindent\textbf{Unified Spatiotemporal Encoding.} We adopt a unified sinusoidal positional encoding $\gamma(\cdot)$ to ensure consistent parameterization across both the geometry deformation and appearance modeling modules. Specifically, given the 3D Gaussian center position \(\mu \in \mathbb{R}^3\) and a scalar time \(t\), the encoding is defined as:
\[
\gamma(p) = \left(\sin(2^k \pi p), \cos(2^k \pi p)\right)_{k=0}^{L-1}, \tag{6}
\]
where \(p\) represents either a spatial coordinate (\(\mu_x, \mu_y, \mu_z\)) or the temporal scalar \(t\), and \(L\) controls the number of frequency bands. The combined input feature for subsequent modules is thus constructed as:\(\left[ \gamma(\mu), \gamma(t) \right].\)
This encoding is consistently shared across both the deformation MLP and the residual appearance MLPs. Empirically, we set \(L=10\) for spatial coordinates and \(L=6\) for temporal encoding in synthetic scenes, while using \(L=10\) for both dimensions in real-world scenarios.

\noindent\textbf{Static Component.} We represent static regions using Gaussian primitives \(\mathcal{N}_s(\mu_s, \Sigma_s)\), whose geometry (\(\mu_s,\Sigma_s\)) is fixed over time. To model temporal variations in appearance, such as illumination changes, we predict a residual to the initial (frozen) appearance parameters:
\[
w_{s,i}(t) = w_{s,i}^{(0)} + \Delta w_{s,i}(t) \tag{8a} 
\]
\[
\Delta w_{s,i}(t) = \mathrm{MLP}_{\mathrm{app}}^{(s)}\big(\,[\gamma(\mu_{s,i}),\, \gamma(t)]\,\big), \tag{8b}
\]
where \(w_{s,i}^{(0)}\) denotes the initial spherical harmonic (SH) coefficients and opacity, which remain fixed, and \(\Delta w_{s,i}(t)\) is the residual predicted by the appearance MLP. Thus, each static Gaussian can temporally adapt its appearance without altering its geometry. We supervise the static reconstruction using a combination of L1 and Structural Similarity (SSIM) losses, computed within a binary static-region mask \(\mathbf{M}\):
\begin{align}
&\mathcal{L}_{\mathrm{static}} =\; \mathcal{L}_1(\hat{I}_s \odot \mathbf{M}, I_{\mathrm{gt}} \odot \mathbf{M}) \notag  \\
&+ \lambda_{\mathrm{ssim}} \cdot (1 -\mathrm{SSIM}(\hat{I}_s \odot \mathbf{M}, I_{\mathrm{gt}} \odot \mathbf{M})), \tag{9}
\end{align}
where \(\hat{I}_s\) is the rendered static image, and \(I_{\mathrm{gt}}\) is the ground-truth reference.

\noindent\textbf{Dynamic Component.} 
A straightforward baseline for modeling dynamic content is to optimize separate Gaussian primitives per timestamp and interpolate post-hoc~\cite{dynamic3dgs}. However, such decoupled modeling lacks temporal coherence and cannot effectively represent continuous motion patterns. Motivated by recent advances~\cite{yang2023deformable3dgs}, we employ a deformation-based formulation. Specifically, we introduce a deformation network, parameterized by \(\theta\), which predicts time-dependent offsets to canonical Gaussian parameters. Given a canonical position \(\mu_d(0)\) and a timestamp \(t\), the deformation network outputs offsets for position, scale, and rotation:
\[
(\delta \mu, \delta r, \delta s) = \mathcal{F}_\theta\big([\gamma(\mu_d(0)), \gamma(t)]\big). \tag{10}
\]

These offsets update the Gaussian's geometry at time \(t\) as:
\begin{align}
\mu_d(t) &= \mu_d(0) + \delta \mu, \tag{11a} \\
\Sigma_d(t) &= \mathbf{A}_d(t)\,\Sigma_d(0)\,\mathbf{A}_d^\top(t) \tag{11b}
\end{align}
where \(\mathbf{A}_d(t)\) is derived from \(\delta r, \delta s\) and governs the anisotropic scaling and orientation. This formulation enables temporally smooth and flexible deformation modeling through shared spatiotemporal encoding.
The dynamic component is supervised via region-specific losses computed over the dynamic mask region \(\mathbf{1} - \mathbf{M}\):
\begin{align}
\small
&\mathcal{L}_{\mathrm{dynamic}} =\;  \mathcal{L}_1(\hat{I}_d, I_{\mathrm{gt}} \odot (\mathbf{1} - \mathbf{M})) \notag \\
& + \lambda_{\mathrm{ssim}} \cdot (1 - \mathrm{SSIM}(\hat{I}_d, I_{\mathrm{gt}} \odot (\mathbf{1} - \mathbf{M}))) ,\tag{12}
\end{align}
where \(\hat{I}_d\) is the dynamically rendered image. This mask-based supervision ensures each Gaussian module receives gradients exclusively from its corresponding visible regions, promoting stable training and effective disentanglement between static and dynamic components.

\noindent\textbf{Remark I. Beyond Prior Decomposition Schemes.}
Recent works~\cite{bard_gs, gaufre, cogs, DeGauss, dynasplat}share the similar sprit of decomposing scenes into dynamic and static parts via auxiliary cues like masks, optical flow, or multi-branch networks. While effective, these methods often suffer from entangled representations, causing motion leakage and unstable training (See \textbf{\textit{Note (I)}} and \textbf{\textit{Note (II)}} in Sec.~\ref{sec:intro}). In contrast, our method performs explicit geometry-level decomposition within the 3D Gaussian representation. Static Gaussians maintain fixed geometry with time-varying appearance, while dynamic Gaussians deform via a shared spatiotemporal encoding. This unified design eliminates the need for dual-stream architectures or occlusion modeling, improving both efficiency and stability. Combined with visibility-driven pruning and depth-aware initialization, our framework achieves disentangled, temporally coherent reconstructions across diverse scenes.

\noindent\textbf{Remark II. Mask-Guided Disentangled Optimization.}
To segment dynamic regions, we employ explicit per-frame binary masks \( \mathbf{M} \in \{0, 1\} \) generated by the open-vocabulary tracker \textit{Track Anything}~\cite{trackanything}. These masks supervise static and dynamic Gaussians separately—using \( \mathbf{M} \) to ensure robust structural disentanglement. Unlike prior methods~\cite{DeGauss, dynasplat, gaufre} that rely on optical flow or learned soft masks for implicit separation, our approach, akin to BARD-GS~\cite{bard_gs}, benefits from externally provided masks for more accurate and temporally consistent guidance. To further enhance disentanglement, we introduce an asymmetric masking strategy: for static regions (Eq.~(9)), both prediction and ground truth are masked by \( \mathbf{M} \), while for dynamic regions (Eq.~(12)), only the ground truth is masked. This preserves occluded static geometry while enabling complete reconstruction of dynamic content for more stable training and improved reconstruction quality.

\subsection{Visibility-Driven Pruning}

In dynamic scene reconstruction, static Gaussians located near view boundaries or occluded regions are often weakly supervised due to limited visibility, leading to unstable optimization and redundancy. To address this, we propose a visibility-driven pruning strategy that quantifies each static Gaussian's long-term contribution over the video sequence.  Unlike SplaTAM~\citep{splatam}, which employs per-frame RGB-D silhouettes for pruning and densification, our method accumulates visibility and opacity across time, ensuring that only Gaussians with persistently low contribution are removed, thereby stabilizing static geometry under monocular supervision.

Let $G_{\mathrm{s}} = \{ G_{s,i}(\mu_{s,i}, \Sigma_{s,i}, w_{s,i}(t)) \}_{i=1}^{N_s}$ denote the set of static Gaussians. For each $G_{s,i} \in G_{\mathrm{s}}$, we define its integrated visibility score as:
\begin{equation}
\bar{V}_{s,i} = \frac{1}{T} \sum_{t=1}^{T} \mathds{1}_{\{ G_{s,i} \mathrm{~rendered ~at~} t \}} \cdot (1 - \alpha_{s,i}(t)) \tag{15}
\end{equation}
where $T$ is the total number of training frames, $\mathcal{V}_t^{(s)} \subseteq G_{\mathrm{s}}$ denotes the subset of visible static Gaussians at time $t$, $\mathds{1}_{\{ G_{s,i} \mathrm{~rendered ~at~} t\}}$ is the indicator function (1 if $G_{s,i}$ is rendered at time $t$), $\alpha_{s,i}(t)$ is the opacity of $G_{s,i}$ at time $t$.

This formulation unifies notation with earlier definitions and reflects both temporal visibility and opacity modulation. We prune low-contribution Gaussians based on $\bar{V}_{s,i}$ thresholds to improve training stability and reduce redundancy.

\begin{figure*}[!ht]
\centering
\includegraphics[width=0.9\textwidth]{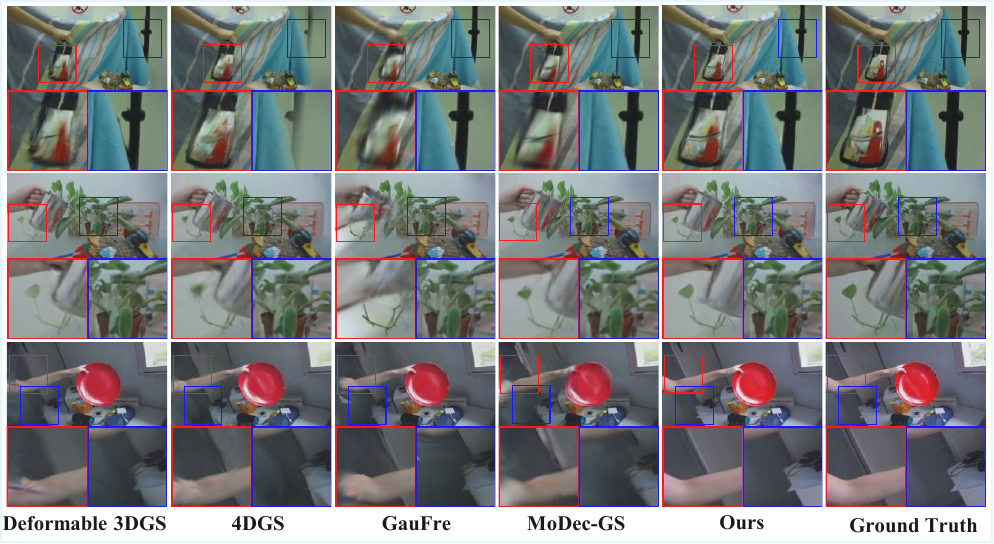} 
\vspace{-5mm}
\caption{\small Qualitative comparison on the NeRF-DS~\cite{nerf_ds} monocular video dataset. Red and blue boxes highlight regions where our method notably improves visual quality.}
\label{fig:sota_NeRF-DS}
\vspace{-3mm}
\end{figure*}

\subsection{Depth-Aware Pretraining}\label{sec:DAP}
3D Gaussian Splatting (3DGS)~\cite{kerbl3Dgaussians} demonstrates that initializing Gaussian primitives using Structure-from-Motion (SfM)~\cite{sfm} point clouds can facilitate effective training~\cite{guo2025implicit}. However, we find that such initializations are often suboptimal for depth-guided reconstruction, frequently leading to geometric inconsistencies. This issue stems from the requirement to align the SfM reconstruction with available depth images by estimating a global scale and offset. In practice, inaccuracies or sparsity in the SfM point cloud can impair this calibration, thereby weakening the efficacy of depth-based regularization. To address this, we introduce a short pretraining stage before the main optimization. This stage refines the initial geometry and improves alignment between the scene structure and depth supervision, resulting in a more reliable point cloud for subsequent learning. Specifically, we regularize the static Gaussian geometry using monocular depth maps through the following loss:
\begin{equation}
    \mathcal{L}_{\mathrm{depth}} = \lambda_{\mathrm{depth}}(t) \cdot \left\| (\hat{D} - D_{\mathrm{gt}}) \odot \mathbf{M} \right\|_1 \tag{14}
\end{equation}
where \( \hat{D} \) denotes the rendered depth map from the static Gaussians, \( D_{\mathrm{gt}} \) is the ground-truth monocular depth map, and \( \mathbf{M} \) is the binary mask identifying static regions. The time-dependent decay factor \( \lambda_{\mathrm{depth}}(t) \) gradually reduces the influence of depth supervision as training progresses. This pretraining stage enhances geometric consistency and stabilizes later optimization. 

\subsection{Training Protocol.}
Our training pipeline follows a two-stage design with an initial \textit{Depth-Aware Pretraining (DAP)} phase to refine SfM-initialized geometry. 

\noindent\textbf{Stage I (S1)} focuses on disentangling static and dynamic components via region-specific supervision using binary masks \( \mathbf{M} \). Static Gaussians are optimized with pixels where \( \mathbf{M} = 1 \), while dynamic Gaussians use the complementary region. To further improve static geometry quality, we introduce a \textit{Visibility-Driven Pruning (VDP)} strategy that removes low-visibility static Gaussians (e.g., near view frustums or edges), thereby mitigating supervision noise and improving geometric stability.

\noindent\textbf{Stage II (S2)} performs joint optimization of both components. Static Gaussians retain fixed geometry and are refined using an \textbf{Appearance Model (APP)}, which predicts view- and time-dependent appearance (e.g., color and opacity) without invoking deformation. Dynamic Gaussians are updated through a learned deformation network conditioned on shared spatiotemporal encodings. During this stage, gradients propagate across both static and dynamic components, allowing mutual refinement. 

Thus, We finally formulate loss function as follows:
\begin{equation}
    \mathcal{L} = \mathcal{L}_{\mathrm{static}} + \mathcal{L}_{\mathrm{dynamic}} + \mathcal{L}_{\mathrm{depth}} \tag{15}
\end{equation}
This two-stage design significantly improves reconstruction fidelity and training stability, as validated in our ablation studies (Table~\ref{tab:Ablation study}), more details can be seen in Figure~\ref{fig:framework}.

\begin{table*}[!ht]
  \centering\small
  \setlength{\tabcolsep}{5pt}
  \resizebox{0.95\textwidth}{!}{
  \begin{tabular}{@{}l  ccc  ccc  ccc  ccc@{}}
    \hline
       & \multicolumn{3}{c}{\textit{as}}
       & \multicolumn{3}{c}{\textit{basin}}
       & \multicolumn{3}{c}{\textit{bell}}
       & \multicolumn{3}{c}{\textit{cup}} \\[4pt]
    \multicolumn{1}{l}{Method}
        & PSNR~$\uparrow$ & SSIM~$\uparrow$ & LPIPS~$\downarrow$ 
        & PSNR~$\uparrow$ & SSIM~$\uparrow$ & LPIPS~$\downarrow$ 
        & PSNR~$\uparrow$ & SSIM~$\uparrow$ & LPIPS~$\downarrow$ 
        & PSNR~$\uparrow$ & SSIM~$\uparrow$ & LPIPS~$\downarrow$ \\
    \hline
    \textbf{HyperNeRF$_\text{SIGGRAPH Asia 2021}$}
      & 25.59       & 0.8567 & 0.1754
      & \Max{20.41}       & \Max{0.8099} & 0.1889
      & 23.06       & 0.7698 & 0.2479
      & 23.98       & 0.8531 & 0.1988 \\
    \textbf{NeRF-DS$_\text{CVPR 2023}$}
      & 25.34       & 0.8679 & 0.1515
      & \Second{20.23}       & \Second{0.8032} & 0.2008
      & 22.57       & 0.7821 & 0.2489
      & \Second{24.51}       & 0.8659 & 0.1668 \\
    \textbf{Deformable\,3DGS$_\text{CVPR 2024}$}$^{*}$
      & \Second{26.03} & \Max{0.8836}      & 0.1351
      & 19.67      & 0.7867 & 0.1498
      & 24.48 & 0.7997   & 0.1822
      & 24.50 & 0.8763   & \Second{0.1472} \\
    \textbf{4DGS$_\text{CVPR 2024}$}$^{*}$
      & 24.77      & 0.8642 & 0.1521
      & 19.36 & 0.7677   & 0.1678
      & 23.16      & 0.8015 & 0.1571
      & 23.88      & 0.8691      & 0.1532 \\
    \textbf{GauF\!re$_\text{WACV 2025}$}$^{*}$
      & \Max{26.05} & 0.8790 & \Second{0.1244}
      & 19.54       & 0.7780      & \Max{0.1222}
      & \Max{25.24} & \Second{0.8130} & \Max{0.1351}
      & 24.04       & 0.8191 & 0.2054 \\
    \textbf{MoDec-GS$_\text{CVPR 2025}$}$^{*}$
      & 24.65   & 0.8538 & 0.1460
      & 19.57   & 0.7787 & 0.1805
      & 22.19   & 0.7562 & 0.2312
      & 24.18   & \Second{0.8798} & 0.2643 \\
    \textbf{Ours}
      & 26.01      & \Second{0.8806} & \Max{0.1031}
      & 19.78 & 0.7885 & \Second{0.1278}
      & \Second{25.55} & \Max{0.8484} & \Second{0.1425}
      & \Max{24.63} & \Max{0.8829} & \Max{0.1112} \\
    \hline
    & \multicolumn{3}{c}{\textit{plate}}
      & \multicolumn{3}{c}{\textit{press}}
      & \multicolumn{3}{c}{\textit{sieve}}
      & \multicolumn{3}{c}{{\textbf{Average}}} \\[4pt]
    \multicolumn{1}{l}{Method}
        & PSNR~$\uparrow$ & SSIM~$\uparrow$ & LPIPS~$\downarrow$ 
        & PSNR~$\uparrow$ & SSIM~$\uparrow$ & LPIPS~$\downarrow$ 
        & PSNR~$\uparrow$ & SSIM~$\uparrow$ & LPIPS~$\downarrow$ 
        & \textbf{PSNR~$\uparrow$} & \textbf{SSIM~$\uparrow$} & \textbf{LPIPS~$\downarrow$} \\
    \hline
    \textbf{HyperNeRF$_\text{SIGGRAPH Asia 2021}$}
      & \Max{21.10}       & 0.7979  & 0.2614
      & 24.59       & 0.8263 & 0.2385
      & 25.41       & 0.8593 & 0.2142
      & 23.44       & 0.8247 & 0.2178 \\
    \textbf{NeRF-DS$_\text{CVPR 2023}$}
      & 19.70       & 0.7813 & 0.2467
      & \Second{25.34}       & \Second{0.8711} & 0.2032
      & 24.99       & \Second{0.8705} & 0.2067
      & 23.24       & 0.8345 & 0.2035 \\  
    \textbf{Deformable\,3DGS$_\text{CVPR 2024}$}$^{*}$
      & 19.88      & \Max{0.8293} & 0.1914
      & 25.32 & \Max{0.8752} & \Max{0.1378}
      & \Second{25.62} & 0.8627 & \Second{0.1206}
      & \Second{23.64}       & \Second{0.8447} & \Second{0.1520} \\
    \textbf{4DGS$_\text{CVPR 2024}$}$^{*}$
      & 18.77 & 0.7891 & \Second{0.1857}
      & 24.81       & 0.8311 & 0.1598
      & 25.16       & 0.8611 & 0.1234
      & 22.84 & 0.8262 & 0.1570 \\
    \textbf{GauF\!re$_\text{WACV 2025}$}$^{*}$
      & 20.00      & 0.8051      & 0.2323
      & 25.05       & 0.8545 & 0.1763
      & 24.88       & 0.8568 & 0.1623
      & 23.54       & 0.8293      & 0.1654 \\
    \textbf{MoDec-GS$_\text{CVPR 2025}$}$^{*}$
      & 18.87      & 0.7306   & 0.2547
      & 22.87      & 0.7296   & 0.2111  
      & 23.48      & 0.7982   & 0.2001 
      & 22.25      & 0.7895   & 0.2125 \\
    \textbf{Ours}
      & \Second{20.34} & \Second{0.8116} & \Max{0.1413}
      & \Max{25.43} & 0.8701     & \Second{0.1498}
      & \Max{26.49} & \Max{0.8753} & \Max{0.1137}
      & \Max{24.03} & \Max{0.8510} & \Max{0.1270} \\
    \hline
  \end{tabular}
  }
  \caption{ Quantitative results on the NeRF-DS~\cite{nerf_ds} dataset. Red and orange cells denote the best and second-best results, respectively. Noted that $(*)$ indicates results reproduced from the authors’ official open-source code under identical fair settings.}
  \vspace{-8mm}
  \label{tab:sota_NeRF-DS}
\end{table*}
\section{Experiment}


\subsection{Experimental Setup}
We implement our method in PyTorch~\cite{pytorch}, building upon the official 3D Gaussian Splatting~\cite{kerbl3Dgaussians} and Deformable 3DGS~\cite{yang2023deformable3dgs} frameworks. In training stage, we first separately optimize static and dynamic components for 30k iterations without the static appearance model; secondly, we jointly train both components with appearance modeling for 40k iterations. A single Adam optimizer~\cite{adam2014method} with \(\beta_1{=}0.9\), \(\beta_2{=}0.999\), and a learning rate decaying exponentially from \(8 \times 10^{-4}\) to \(1.6 \times 10^{-6}\) is used. Both deformation and appearance MLPs share architecture and schedule. Visibility-driven pruning removes rarely rendered Gaussians to improve training efficiency. For depth-supervised scenes, a short pretraining stage aligns SfM point clouds with depth maps before applying depth regularization. All experiments are conducted on a single NVIDIA 5090 GPU (32GB). Training a typical sequence takes around 1 hour, with peak memory usage below 30 GB, and inference runs at 200 FPS for a resolution of 480×270. We evaluate our method on three datasets: (1) \textbf{iPhone}~\cite{iphone}, featuring 14 real-world scenes (4180 frames at \(720 \times 480\)) with handheld motion and diverse dynamics; (2) \textbf{NeRF-DS}~\cite{nerf_ds}, a monocular dataset containing specular objects and challenging motion; and (3) \textbf{HyperNeRF}~\cite{park2021hypernerf}, which includes complex dynamic scenes captured in real-world environments. Following standard protocols, we report PSNR, SSIM, and LPIPS~\cite{lpips}.
\begin{figure}[!htb]
\centering
\includegraphics[width=0.95\linewidth, trim=0mm 0mm 3mm 1mm, clip]{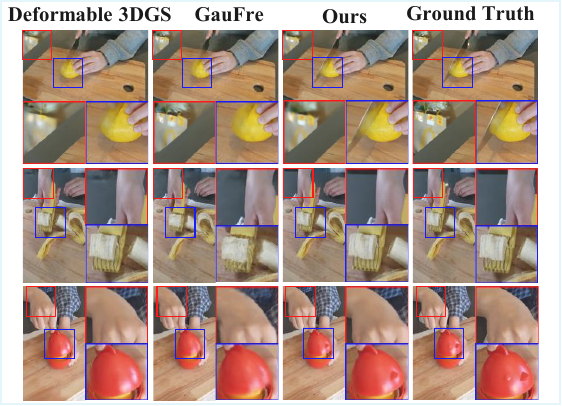} 
\vspace{-5mm}
\caption{\small Qualitative results on HyperNeRF~\cite{nerf_ds}.}
\label{fig:sota_HyperNeRF}
\vspace{-3mm}
\end{figure}

\subsection{Compared with SOTA Results}
\noindent\textbf{Quantitative Comparisons.}
We compare our approach with Deformable 3DGS~\cite{yang2023deformable3dgs}, 4DGS~\cite{Wu_2024_CVPR}, HyperNeRF~\cite{park2021hypernerf}, NeRF-DS~\cite{nerf_ds}, GauFre~\cite{gaufre}, MoDec-GS~\cite{modecgs}. As summarized in Table~\ref{tab:sota_NeRF-DS}, our method consistently achieves the best or second-best performance on the NeRF-DS dataset, demonstrating robustness under challenging dynamics and lighting. Averaged over all metrics, it surpasses competing baselines, indicating superior reconstruction capability. Table~\ref{tab2} further reports evaluations on the HyperNeRF and iPhone datasets, where our method remains highly competitive against Deformable 3DGS and GauFre. Noted that the iPhone dataset is inherently challenging due to handheld capture, rolling-shutter distortions, and unstable exposure, and prior dynamic Gaussian-splatting methods also report universally low PSNR on this benchmark. Moreover, we further include three recent dynamic Gaussian-splatting methods 4DGS~\cite{Wu_2024_CVPR}, 4DGS2~\cite{yang2023gs4d}, and MoDec-GS~\cite{modecgs} to provide a more complete evaluation on the HyperNeRF~\cite{park2021hypernerf} dataset. These works represent diverse strategies for modeling spatiotemporal variation beyond traditional 3D Gaussian formulations. The results from Table~\ref{tab:hypernerf_extended} demonstrate that SplitGaussian remains highly competitive among a diverse set of recent dynamic Gaussian-splatting methods. Despite the varied modeling assumptions of these baselines, our approach achieves consistently strong performance across all metrics, particularly in perceptual reconstruction quality. This highlights the robustness of our geometry–appearance decomposition and confirms that SplitGaussian maintains stable reconstruction performance even when compared with more complex 4D or motion-decomposition formulations. These results collectively validate the generalization of our approach across diverse real-world settings involving non-rigid motion and handheld camera trajectories.

\begin{figure}[htbp]
\centering
\includegraphics[width=0.95\linewidth, trim=0 0 0 4.2cm, clip]{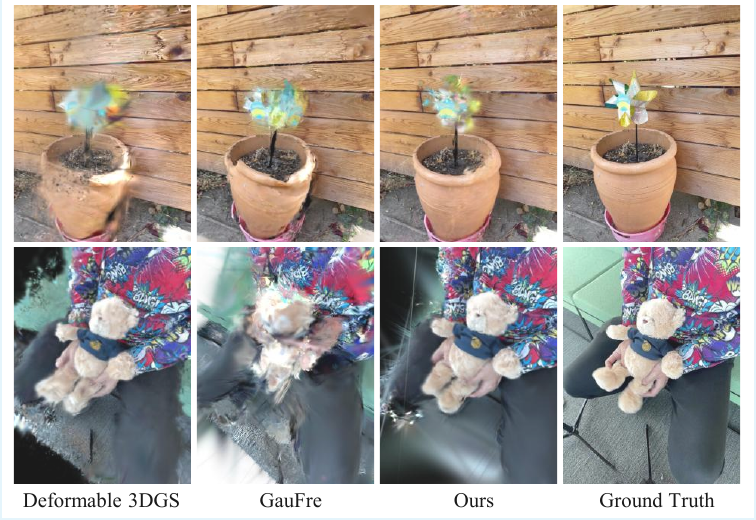} 
\vspace{-4mm}
\caption{\small Qualitative comparison on the iphone~\cite{iphone} dataset. }
\vspace{-3mm}
\label{fig:sota_iphone}
\end{figure}

\noindent\textbf{Qualitative Comparisons.}
We qualitatively evaluate our method on NeRF-DS~\cite{nerf_ds} and HyperNeRF~\cite{park2021hypernerf}, as shown in Figure~\ref{fig:sota_NeRF-DS} and Figure~\ref{fig:sota_HyperNeRF}. These datasets cover dynamic motions, complex lighting, and varying camera paths. Our method achieves high-fidelity results with sharp details, temporally coherent motion, and clean static regions. In particular, it maintains geometric stability and appearance consistency even where dynamic and static elements interact, outperforming existing baselines. This highlights the effectiveness of our decomposition and unified spatiotemporal modeling for robust and perceptually plausible dynamic scene reconstruction. In addition, we present qualitative comparisons on the iPhone~\cite{iphone} dataset in Figure~\ref{fig:sota_iphone}. As a result, baseline methods such as Deformable 3DGS~\cite{yang2023deformable3dgs} and GauFRe~\cite{gaufre} tend to exhibit severe geometric drift, motion over-smoothing, and texture collapse. On the contrary, our SplitGaussian method delivers significantly more stable static geometry and cleaner static–dynamic separation. While dynamic object sharpness remains below ground truth, it substantially reduces motion leakage and preserves consistent appearance across time. Performance is further limited by the iPhone dataset’s multiple fixed viewpoints, as our training uses only one viewpoint per sequence, restricting parallax cues and geometric supervision and thus lowering accuracy relative to moving-camera datasets. Despite these constraints, SplitGaussian maintains competitive reconstruction quality and outperforms existing dynamic Gaussian-splatting methods in temporal consistency.

\begin{table}
\centering
\small
\setlength{\tabcolsep}{4pt}
\resizebox{0.95\linewidth}{!}{
\begin{tabular}{l ccc}
\hline
\multicolumn{4}{c}{HyperNeRF} \\
\hline
Method & PSNR~$\uparrow$ & SSIM~$\uparrow$ & LPIPS~$\downarrow$ \\
\hline
\textbf{Deformable 3DGS$_\text{CVPR 2024}$} $^{*}$ 
  & \Second{24.57} 
  & \Max{0.7641} 
  & 0.2439 \\
\textbf{GauF\!re$_\text{WACV 2025}$}$^{*}$ 
  & 23.59 
  & 0.7486 
  & \Second{0.2416} \\
\textbf{Ours} 
  & \Max{24.61} 
  & \Second{0.7626} 
  & \Max{0.2398} \\
\hline
\multicolumn{4}{c}{iPhone} \\
\hline
Method & PSNR~$\uparrow$ & SSIM~$\uparrow$ & LPIPS~$\downarrow$ \\
\hline
\textbf{Deformable 3DGS$_\text{CVPR 2024}$}$^{*}$ 
  & 12.56 
  & 0.2902 
  & \Max{0.5896} \\
\textbf{GauF\!re$_\text{WACV 2025}$}$^{*}$ 
  & \Second{13.27} 
  & \Second{0.3382} 
  & 0.6206 \\
\textbf{Ours} 
  & \Max{13.53} 
  & \Max{0.3391} 
  & \Second{0.6205} \\
\hline
\end{tabular}}

\caption{\small Results on HyperNeRF~\cite{park2021hypernerf} and iPhone~\cite{iphone} datasets.
}
\label{tab2}
\vspace{-5mm}
\end{table}

\begin{table}
\centering
\resizebox{0.95\linewidth}{!}{
\begin{tabular}{lccc}
\toprule
\textbf{Method} 
& \textbf{PSNR} $\uparrow$ 
& \textbf{SSIM} $\uparrow$ 
& \textbf{LPIPS} $\downarrow$ \\
\midrule
\textbf{Deformable 3DGS$_\text{CVPR 2024}$} $^{*}$ & 24.57 & \Max{0.7641} & 0.2439 \\
\textbf{4DGS$_\text{CVPR 2024}$} $^{*}$        & \Second{24.82} & 0.7569 & 0.2472 \\
\textbf{4DGS2$_\text{ILCR 2024}$} $^{*}$           & 24.47 & 0.7426 & 0.2508 \\
\textbf{GauFRe$_\text{WACV 2025}$} $^{*}$        & 23.59 & 0.7486 & \Second{0.2416} \\
\textbf{MoDec-GS$_\text{CVPR 2025}$} $^{*}$       & \Max{24.87} & 0.7474 & 0.2588 \\
\textbf{Ours}               & 24.61 & \Second{0.7626} & \Max{0.2398} \\
\bottomrule
\end{tabular}}
\caption{\small Extended quantitative comparison on the HyperNeRF~\cite{park2021hypernerf} dataset including additional dynamic Gaussian-splatting baselines.}
\label{tab:hypernerf_extended}
\vspace{-8mm}
\end{table}

\subsection{Ablation Study}

\begin{figure*}[htbp]
\centering
\includegraphics[width=0.95\textwidth]{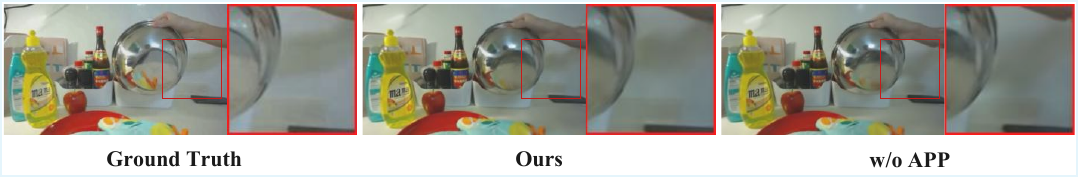} 
\vspace{-5mm}
\caption{\small Ablation on Appearance Modeling (APP). Comparison of reconstruction results with and without the appearance module. Ours exhibits better alignment and fewer artifacts in static regions.}
\vspace{-3mm}
\label{fig:Ablation_APP}
\end{figure*}

\noindent\textbf{Visual Geometry Decomposition Matters.}
We conduct an ablation study on the NeRF-DS dataset to quantify each component in the \textbf{SplitGaussian} framework, as summarized in Tab.~\ref{tab:Ablation study}.  
Starting from the baseline (a) S1, which separately trains static and dynamic branches, we observe modest reconstruction quality. Adding S2 in (b) introduces joint optimization between both branches, leading to consistent gains of +0.94 dB in PSNR, +0.0145 in SSIM, and a LPIPS drop of 0.0111, indicating the benefit of mutual spatiotemporal supervision. The introduction of the appearance modeling module in (c) further reduces perceptual distortion (LPIPS $\downarrow$ 0.0227), suggesting improved handling of time-varying appearance. In (d), depth-aware pretraining facilitates geometric alignment, contributing an additional +0.42 dB in PSNR and a 0.0163 reduction in LPIPS. Finally, incorporating visibility-driven pruning in (e) yields the best overall performance, achieving a PSNR of 24.03, SSIM of 0.8505, and LPIPS of 0.1274. These results indicate that each module brings notable improvement, and the full model provides the most stable and perceptually accurate reconstruction.

\begin{table}
\centering
\resizebox{0.95\linewidth}{!}{
\begin{tabular}{lccc}
\hline
Variant & PSNR~$\uparrow$ & SSIM~$\uparrow$ & LPIPS~$\downarrow$ \\
\hline
(a) S1(baseline)                         & 22.41 & 0.8268 & 0.1843 \\
(b) S1+S2                      & 23.35 & 0.8413 & 0.1732 \\
(c) S1+S2+APP                  & 23.36 & 0.8419 & 0.1505 \\
(d) S1+S2+APP+DAP               & 23.78 & 0.8457 & 0.1342 \\
(e) (d)+VDP(\text{full})               & 24.03 & 0.8505 & 0.1274 \\
\hline
\end{tabular}}
\caption{\small Ablation study on individual components of SplitGaussian on the NeRF-DS dataset.}
\label{tab:Ablation study}
\vspace{-10mm}
\end{table}

\noindent\textbf{Appearance Modeling Improves Static Reconstruction.}
Figure~\ref{fig:Ablation_APP} illustrates reconstructions \textbf{w/} and \textbf{w/o} APP. Removing the appearance module results in noticeable degradation in static regions, especially under subtle lighting changes, leading to artifacts.
A closer examination reveals that the model with APP can accurately reconstruct fine-grained illumination effects—such as the soft shadow on the static paper surface—while the version without APP fails to capture this detail, producing over-smoothed or inconsistent shading.
This highlights the necessity of temporally adaptive appearance modeling for static Gaussians to ensure consistent visual quality across time.

\begin{figure}[htbp]
\centering
\includegraphics[width=\columnwidth]{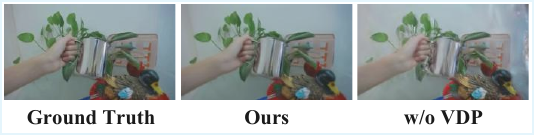}
\vspace{-6mm}
\caption{\small Removing \textbf{VDP} causes bright boundary artifacts from undertrained Gaussians in sparsely observed regions.}
\label{fig:Ablation_VDP}
\vspace{-3mm}
\end{figure}

\noindent\textbf{VDP Improves Peripheral Realism.}
Figure~\ref{fig:Ablation_VDP} illustrates the impact of removing the Visibility-Driven Pruning (VDP) module. Without VDP, static Gaussians located near the periphery of training views, where visibility is sparse, receive inadequate optimization. As a result, these insufficiently supervised Gaussians accumulate during densification and exhibit artificially elevated opacities. This leads to prominent residual artifacts, particularly along image boundaries, which degrade rendering quality and visual coherence. In contrast, our full method incorporates visibility-aware filtering to suppress low-visibility Gaussians early in training, effectively reducing peripheral noise and producing cleaner, artifact-free 

\begin{figure}[htbp]
\centering
\includegraphics[width=\columnwidth]{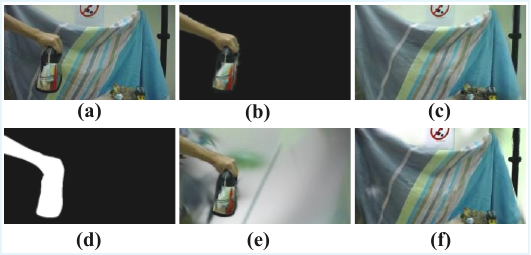} 
\vspace{-8mm}
\caption{\small (a) Ground truth. (b) Our dynamic reconstruction. (c) Our static reconstruction. (d) Learned dynamic mask shared by both methods. (e) Dynamic result from GauFRe~\cite{gaufre}. (f) Static result from GauFRe~\cite{gaufre}.}
\label{figGau}
\vspace{-5mm}
\end{figure}

\noindent\textbf{Beyond Mask-Guided Approaches.}
Figure~\ref{figGau} presents a visual comparison between our method and GauFRe~\cite{gaufre}. GauFRe effectively integrates occlusion reasoning and mask-based decomposition, enabling basic separation of dynamic and static regions. However, as shown in (e)(f), it occasionally produces blending and structural artifacts under complex motion. In contrast, our approach (b)(c), guided by a learned dynamic mask (d), achieves clearer decomposition in both regions, suggesting enhanced robustness in dynamic-static separation.

\begin{figure}[htbp]
\centering
\includegraphics[width=\columnwidth]{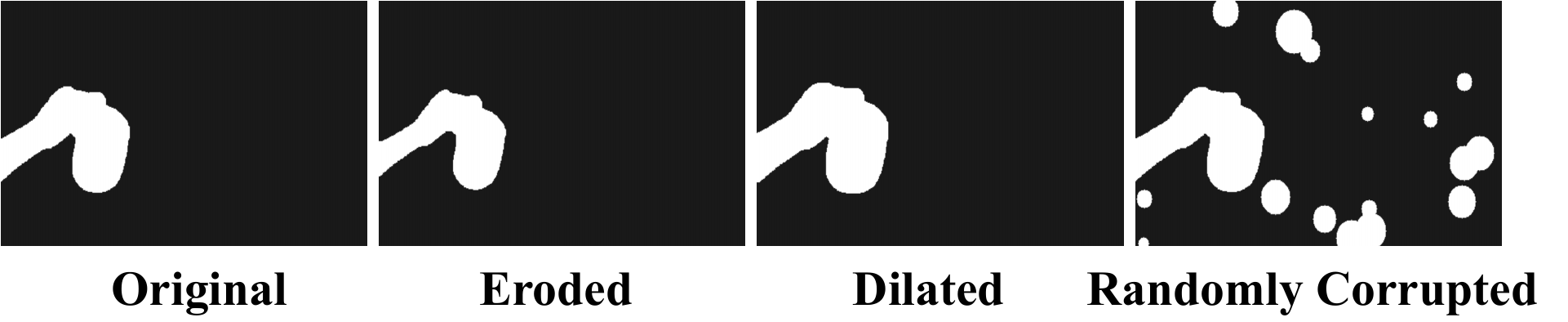} 
\vspace{-8mm}
\caption{\small Illustration of the mask perturbation experiment. From left to right: Original Mask, Eroded Mask, Dilated Mask, and Randomly Corrupted Mask.}
\vspace{-3mm}
\label{fig:mask}
\end{figure}

\noindent\textbf{Robustness Analysis under Mask Perturbations.}
To evaluate SplitGaussian under imperfect dynamic-mask supervision, we conduct controlled experiments on the NeRF-DS dataset~\cite{nerf_ds} by systematically perturbing the input binary masks under different degradation settings. As illustrated in Figure~\ref{fig:mask}, we construct four variants of the mask: (i) the \textbf{Original Mask} generated by the Track Anything framework~\cite{trackanything}, (ii) an \textbf{Eroded Mask} produced via morphological erosion, which shrinks the dynamic region by removing boundary pixels, (iii) a \textbf{Dilated Mask} produced via morphological dilation, which enlarges the dynamic region beyond its true extent, and (iv) a \textbf{Randomly Corrupted Mask} in which random holes are inserted to mimic noise, missing areas, or unstable tracking.

As shown in Table~\ref{tab:mask_ablation}, our model achieves the highest reconstruction quality when using the Original Mask, which reflects the upper bound performance under accurate mask supervision. The Randomly Corrupted Mask yields results close to the baseline, demonstrating that SplitGaussian is resilient even when the mask is fragmented or contains missing regions. This finding indicates that the model does not overfit to local mask noise and can maintain coherent reconstruction despite irregular supervision artifacts. Overall, the small performance variations across perturbed masks confirm that SplitGaussian is robust to moderate mask inaccuracies, further validating the effectiveness of the proposed static–dynamic decomposition and its reduced dependence on precise pixel-level segmentation.

\begin{table}[htbp]
\centering
\resizebox{\linewidth}{!}{
\begin{tabular}{lccc}
\toprule
\textbf{Mask Variant} 
& \textbf{PSNR} $\uparrow$ 
& \textbf{SSIM} $\uparrow$ 
& \textbf{LPIPS} $\downarrow$ \\
\midrule
Eroded Mask   & 23.72 & 0.8302 & 0.1458 \\
Dilated Mask  & 23.50 & 0.8332 & 0.1571 \\
Randomly Corrupted Mask & 23.64 & 0.8408 & 0.1498 \\
Original Mask & 24.03 & 0.8510 & 0.1270 \\
\bottomrule
\end{tabular}}
\caption{\small Quantitative results of mask perturbation experiments on the NeRF-DS~\cite{nerf_ds} dataset. 
We report reconstruction quality using PSNR/SSIM/LPIPS.}
\label{tab:mask_ablation}
\vspace{-5mm}
\end{table}

\section{Limitations and Conclusions}



Although SplitGaussian achieves SOTA performance in dynamic scene reconstruction, separately optimizing static and dynamic components before joint appearance modeling introduces additional computational overhead compared to end-to-end methods. Nevertheless, the proposed decomposition of geometry and appearance, combined with unified spatiotemporal encoding, visibility-driven pruning, and depth-aware regularization, enables temporally coherent and photorealistic reconstruction. Extensive experiments confirm its robustness and generalization across diverse dynamic scenes.

\section*{Acknowledgments}
This work has been supported by the National Natural Science Foundation of China (Grant No. 62472139, 62502144, 62502142). The computation is completed on the HPC Platform of Hefei University of Technology.


\bibliographystyle{ACMMM2026_SplitGaussian}
\bibliography{ACMMM2026_SplitGaussian}


\end{document}